\documentclass[letterpaper]{article} 
\usepackage{aaai2026}  
\usepackage{times}  
\usepackage{helvet}  
\usepackage{courier}  
\usepackage[hyphens]{url}  
\usepackage{amsfonts}
\usepackage{graphicx} 
\usepackage{natbib}  
\usepackage{caption} 
\usepackage[cmyk]{xcolor}
\usepackage{algorithm}
\usepackage{algorithmic}

\usepackage{newfloat}
\usepackage{listings}
\DeclareCaptionStyle{ruled}{labelfont=normalfont,labelsep=colon,strut=off} 
\floatstyle{ruled}
\newfloat{listing}{tb}{lst}{}
\floatname{listing}{Listing}
\usepackage{microtype}
\usepackage{url}
\usepackage{booktabs}
\usepackage{xspace}
\usepackage{lineno}
\usepackage{adjustbox}
\usepackage{multirow}
\usepackage{todonotes}
\usepackage[table]{xcolor} 

\newcommand{\ourtechnique}{\textsc{XITE}\xspace}
\definecolor{mygreen}{cmyk}{0.3, 0, 0.6, 0}
\definecolor{lightgreen}{cmyk}{0.1, 0, 0.1, 0}

\title{\ourtechnique: Cross-lingual Interpolation for Transfer using Embeddings}

\author{
    Barah Fazili,
    Preethi Jyothi
}

\affiliations{
    Indian Institute of Technology Bombay, India\\
    \{barah, pjyothi\}@cse.iitb.ac.in
}

\usepackage{bibentry}

\begin{document}

\maketitle

\begin{abstract}
Facilitating cross-lingual transfer in multilingual language models remains a critical challenge. Towards this goal, we propose an embedding-based data augmentation technique called \ourtechnique. We start with unlabeled text from a low-resource target language, identify an English counterpart in a task-specific training corpus using embedding-based similarities and adopt its label. Next, we perform a simple interpolation of the source and target embeddings to create synthetic data for task-specific fine-tuning. Projecting the target text into a language-rich subspace using linear discriminant analysis (LDA), prior to interpolation, further boosts performance. Our cross-lingual embedding-based augmentation technique \ourtechnique yields significant improvements of up to 35.91\% for sentiment analysis and up to 81.16\% for natural language inference, using XLM-R, for a diverse set of target languages including Korean, Arabic, Urdu and Hindi. Apart from boosting cross-lingual transfer, adaptation using \ourtechnique also safeguards against forgetting and maintains task performance on the high-resource language. 
\end{abstract}

\section{Introduction}
Multilingual language models (MLMs), such as XLM-RoBERTa and multilingual BERT, have demonstrated remarkable cross-lingual transfer capabilities, often achieving strong zero-shot performance on downstream tasks. These models benefit from shared multilingual embedding spaces, enabling transfer from high-resource languages (typically English) to low-resource languages without explicit cross-lingual supervision. However, despite their success, the effectiveness of this transfer often depends on the degree of alignment between the source and target representations~\citep{deshpande-etal-2022-bert, dufter-schutze-2020-identifying}. Misalignment, caused by differences in linguistic properties, lexical variations, or domain shifts, can reduce transfer efficiency, particularly for low-resource languages.

We present a new approach \ourtechnique (Cross-lingual (\textbf{X}) \textbf{I}nterpolation for \textbf{T}ransfer using \textbf{E}mbeddings) to enhance cross-lingual transfer by interpolating embeddings between source and target languages. The intuition behind our method is that mixing source and target representations can help alleviate misalignments between their respective embedding spaces, thus encouraging the model to generalize better across languages, unlike prior approaches that rely on explicit alignment via projections or adversarial learning~\citep{cao2020multilingualalignmentcontextualword,chen-etal-2018-adversarial}.

\ourtechnique starts with unlabeled task-specific data in a target language. These text instances are mapped to instances in a labeled source language corpus and source labels are projected onto the target text. Next, we investigate two main interpolation strategies. First, we simply add the embeddings of target language inputs with those of their mapped source counterparts during task finetuning, effectively mixing representations to balance source-task information and information encoded in the target text. In another approach, target embeddings are first projected onto a language-separability basis. This basis is derived by applying Linear Discriminant Analysis (LDA) to sentences from both languages, using language identity as the class label during dimensionality reduction. We interpolate these LDA-driven embeddings for the target sentences (capturing language-specific information) with the embeddings from the source sentences (capturing meaning-specific information). By isolating language-specific features in the target text through projection onto axes that maximize language separability, we study whether emphasizing the (presumably) semantically richer source embeddings during interpolation can further enhance cross-lingual transfer.

We evaluate \ourtechnique on sentiment analysis (SA) and natural language inference (NLI) -- two key tasks in cross-lingual natural language understanding -- using strong encoder-only LMs.%
\footnote{Our choice of encoder-only LMs is justified in detail in Section~\ref{sec:modeltraining}}.
We demonstrate that XITE outperforms standard cross-lingual baselines on multiple target languages 
including Hindi, Arabic, Urdu and Korean. Furthermore, we show that interpolation consistently maintains its performance on the source language as well; in contrast, simple finetuning over the target text is prone to catastrophic forgetting.

Our contributions can be summarized as follows:
\begin{itemize}
    \item We propose a new embedding-based interpolation framework \ourtechnique that bridges the source-target representation gap, enhancing cross-lingual transfer for downstream tasks.
    \item We also introduce a novel LDA based interpolation strategy to highlight the linguistic features from target language text, while drawing semantic information primarily from semantically similar source text.
    \item We demonstrate the effectiveness of \ourtechnique through extensive experiments on SA and NLI tasks across multiple target languages, showing consistent gains over baseline models.
\end{itemize}


\section{Related Work}
Cross-lingual transfer learning focuses on transferring knowledge from source languages to target languages. Prior cross-lingual transfer techniques can be broadly categorized into three types: instance transfer, parameter transfer, and feature transfer~\cite{5288526}. Instance transfer involves translating data or augmenting cross-lingual data, as seen in methods like translate-train and translate-test~\citep{artetxe-etal-2020-translation}.Parameter transfer focuses on learning model architectures that share parameters between languages, facilitating knowledge transfer while feature transfer aims to learn language-invariant features through techniques such as adversarial networks or re-alignment methods. \citet{chen2019multisourcecrosslingualmodeltransfer} proposed a model that leverages adversarial networks to learn language-invariant features, enabling effective cross-lingual transfer by dynamically exploiting similarities between the target language and each individual source language. 

Mixup, introduced by ~\citet{zhang2018mixupempiricalriskminimization}, on the other hand is a regularization technique that trains models on linear interpolations of input data and their corresponding labels, effectively improving model robustness and generalization.
\citet{guo2019augmentingdatamixupsentence} adapted mixup for sentence classification by performing interpolation on word embeddings and sentence embeddings using CNN and LSTM models.
\citet{sun-etal-2020-mixup} introduced the Mixup-Transformer, integrating mixup into transformer-based architectures for various NLP tasks. This approach involves interpolating in the embedding space rather than raw text, allowing for effective data augmentation with significant performance improvements. However, their work is limited to English, with no extension to multilingual or cross-lingual settings.

Recent work has further adapted the mixup technique, originally proposed for images, to better suit multilingual contexts. Manifold Mixup, proposed by ~\citet{verma2019manifoldmixupbetterrepresentations}, extends the idea of mixup by performing interpolation in the latent space via linear combinations of hidden states. While these methods have shown success in monolingual settings, their application in cross-lingual contexts presents new challenges. X-MIXUP~\cite{yang2022enhancingcrosslingualtransfermanifold}  addresses these by mixing representations of source and target languages during training and inference, enabling source-aware cross-attention training on both source and translated target text. This approach facilitates the extraction of target-related information from source hidden states, enhancing cross-lingual transfer performance. However, it relies on translation of labeled source data into the target language during training and vice versa at inference, synthesizing predictions from both the source and target sequences by averaging their predicted probability distributions. In contrast, our approach avoids any reliance on translation, making it more scalable to low-resource languages.

\citet{tang2022robustunsupervisedcrosslingualword} proposed an unsupervised method that introduces intermediate pseudo-language spaces through linear interpolation. 
\citet{xie-etal-2022-discovering} identify a low-rank subspace within multilingual embeddings that encodes language-specific signals. By projecting embeddings into the null space of this subspace using singular value decomposition (SVD) on multiple monolingual corpora across the languages, they suppress language identity while retaining semantic information, thereby improving performance on cross-lingual tasks. While null space projection explicitly aims to eliminate language-specific information, our approach takes a different path. We explore two strategies: directly interpolating the original embeddings of source and target text, and isolating language-specific features from the target text before mixing them with the embeddings of semantically similar source language text. This approach aims to balance cross-lingual alignment while preserving meaningful linguistic nuances in the target language.

\citet{zhang-etal-2020-seqmix} propose token-level interpolation for sequence labeling tasks such as named entity recognition. It selects instance pairs using active learning heuristics (e.g., least confidence, normalized token entropy) and forms synthetic samples by mixing token embeddings. These interpolated embeddings are then mapped back to valid tokens using contextual similarity. The method operates purely within the monolingual domain and relies on label-aware constraints to ensure meaningful interpolation. Similarly, \citet{chen2020mixtextlinguisticallyinformedinterpolationhidden}  performs interpolation in the feature space between ``easy-to-learn" and ``ambiguous" examples based on training dynamics. 
\citet{park-caragea-2022-data} introduces layer-wise hidden state interpolation in pretrained language models like BERT. Interpolation is done at intermediate layers, and a KL divergence loss is used to align the soft labels of the interpolated samples with the model's predictions. This method focuses on semi-supervised text classification and operates solely in monolingual contexts. In contrast to the above approaches, our work targets cross-lingual generalization, where interpolation must bridge structural and semantic mismatches across languages. We address the challenges of aligning and interpolating representations from linguistically diverse languages, which is fundamentally different from interpolating within a single language.
\section{Methodology}
\label{sec:mapping}
\ourtechnique comprises three main steps that are described in detail below.

\subsection{Mapping target text to source text}
\label{sec:mapping}
For unlabeled text in a target language, we choose any labeled corpus for the task in a resource-rich source language (e.g., English) and map each target-language instance to its closest source-language counterpart by computing cosine similarities between their embeddings\footnote{Unless specified otherwise, text representations are averaged contextualized token embeddings from a transformer-based encoder.}. For this mapping, we create an index over the source (English) corpus, and for each target-language instance, we retrieve the top-$m$ most similar English instances based on cosine similarities from model embeddings. The selected source data points are further used for both label projection and embedding interpolation, as discussed below.

\subsection{Deriving language-rich basis}
\label{sec:deriv-basis}
We hypothesize that projecting the embeddings over language-specific features in target-language text before interpolation can improve cross-lingual transfer by preserving key linguistic information. To this end, we adopt a modified Linear Discriminant Analysis (LDA) approach, inspired by prior work~\cite{shah-etal-2024-correlations}. 

For language separability in multilingual representation spaces, we construct a new orthonormal basis 
$V$, where the axes are sorted based on their language-separability. We start by applying linear discriminant analysis (LDA) to sentence representations (obtained by averaging token embeddings across the sentence)  from a corpus composed of sentences from the two languages (English and a target language), using language as the label. The first LDA axis $v_0$, which maximally separates the languages, initializes the basis:$V = v_0 \in \mathbb{R}^{d \times 1}$.

\begin{figure}[h]
  \centering
  \includegraphics[width=1\linewidth, trim=90 40 60 40, clip]{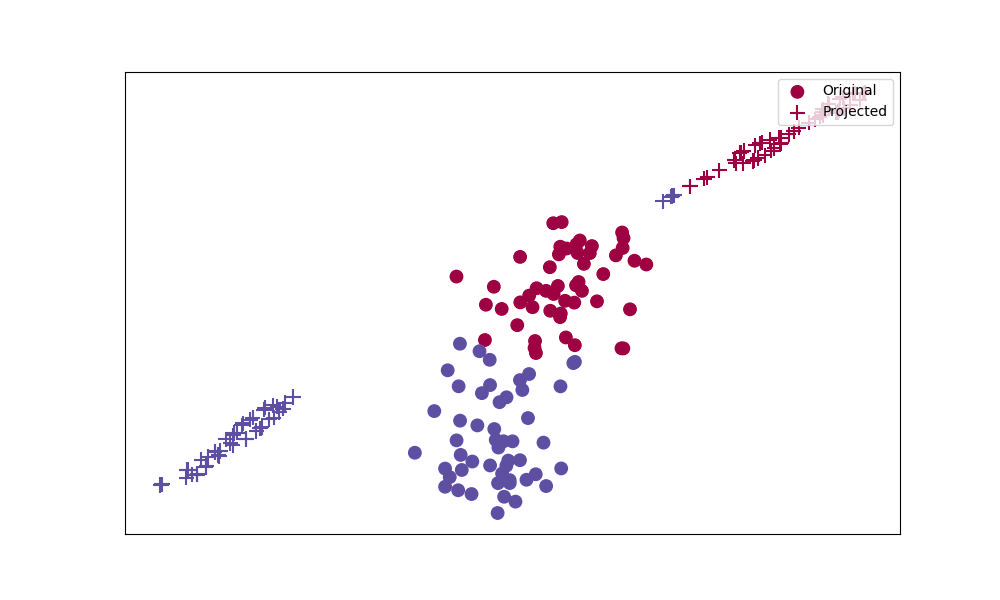} 
  \caption{t-SNE plot showing sentences in English and Arabic, before and after projection using the basis vectors derived in Section~\ref{sec:deriv-basis}.}
  \label{fig:tsne}
\end{figure}

Next, we iteratively perform the following steps:
\begin{enumerate}
\item \textit{Projection and Subtraction}: We project all sentence representations $X\in \mathbb{R}^{d \times n}$ onto the current set of language-separable axes $V$ and subtract the projections from the original sentence representations: $X - VV^T X$. This step aims to remove the language-separable components identified by $V$, by setting those dimensions to fixed values across all sentence representations.
\item \textit{Finding the Next Axis}: We apply LDA on the adjusted representations to identify the next highest language-separable axis 
$v_i$. We orthogonalize $v_i$ relative to the current axes in $V$, and update the basis: $V \leftarrow [V, v_i]$.
\item \textit{Iteration}: We repeat this process $k$ times to get $k$ orthogonal axes.
\item \textit{Projection}: Finally, we project the text $X$ along this basis matrix (containing $k$ orthogonal axes) and project it back to the original model embedding size $d$ (for easy interpolation).
\end{enumerate}
Based on this iterative process, the resulting axes in the basis are more language-separable, as they represent dimensions that maximize language distinctions. (If this process was continued until $k=d$ i.e. the original embedding size and the latter axes were taken instead, we would be extracting language neutral axes.) Figure \ref{fig:tsne}
shows a t-SNE
plot of text embeddings for fifty sentences each from Arabic and English before and after projection. After projection with our basis, the language clusters (shown in two different colors) become tighter and the separation between language clusters is also visually improved.

\subsection{Interpolation during finetuning}
\begin{figure}[h]
  \centering
  \includegraphics[width=1\linewidth]{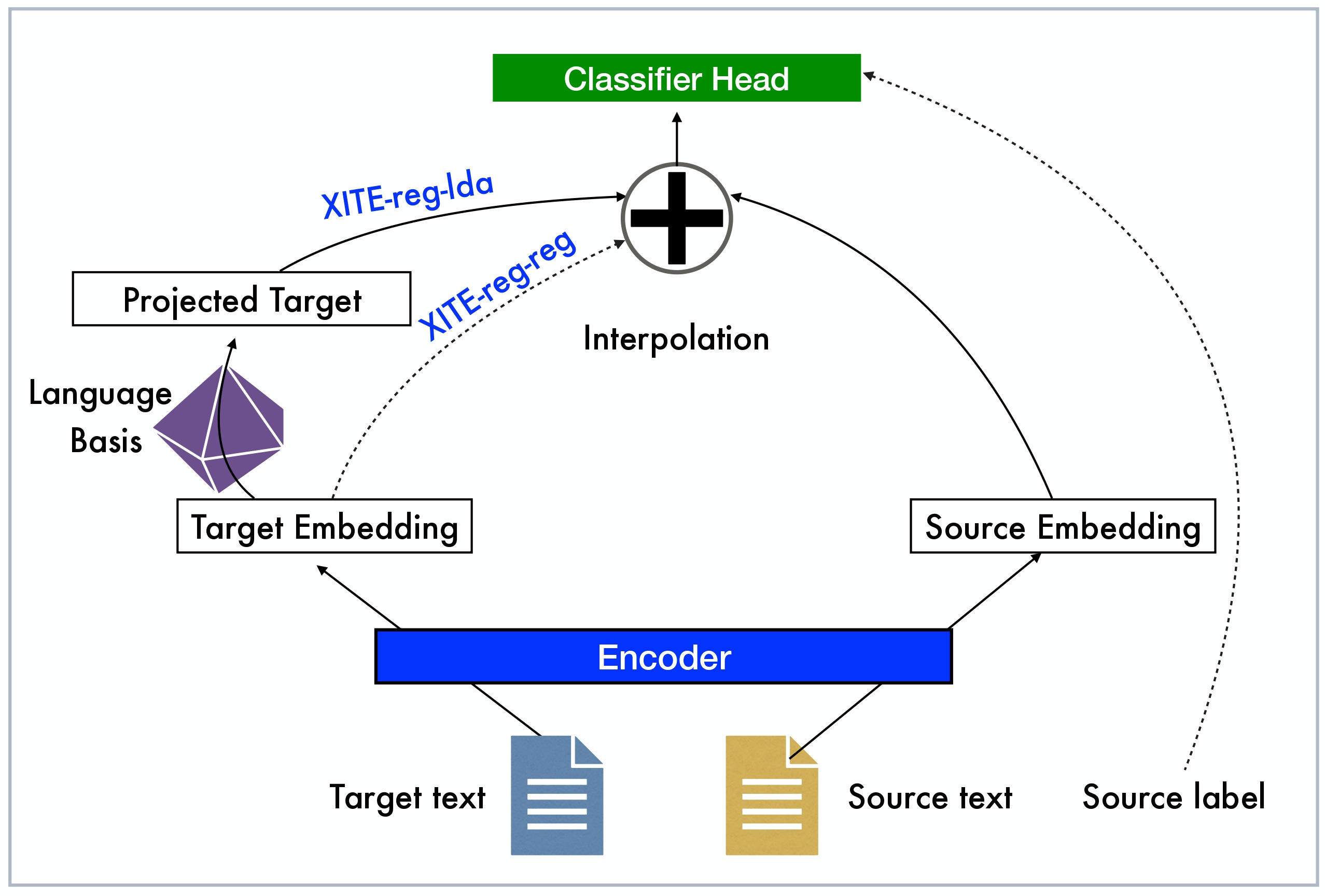} 
  \caption{Schematic overview of interpolation step in \ourtechnique: embeddings are combined either by direct addition (XITE-reg-reg) or after projecting the target onto a language-rich basis (XITE-reg-lda) before feeding the interpolated representation into the classifier.}
  \label{fig:design}
\end{figure}

The final step of \ourtechnique involves performing a simple interpolation between the source and target language text embeddings. This is achieved via dual-encoder training, where both target and corresponding source embeddings are derived from the encoder. The target embedding is first projected onto the language-separability basis (derived in Section~\ref{sec:deriv-basis}) for the target-source pair and then projected back to its original size to maintain compatibility with the classifier architecture. The resulting projected target embedding is then added to the source embedding, and this mixed embedding is fed into the classifier/prediction head. Along with the interpolated embedding, the projected label from the source text is directly used during fine-tuning for the classification task. 
Figure \ref{fig:design} shows an illustration of both interpolation without projection (\ourtechnique--\texttt{reg-reg}) and interpolation after projection (\ourtechnique-\texttt{reg-lda}). 

\section{Experiments}
\label{sec:expts}
We experiment with the Sentiment Analysis (SA) task for Korean, Arabic and Hindi and the Natural Language Inference (NLI) task for Korean, Arabic, Hindi and Urdu.\footnote{Mean accuracies are reported for both classification tasks.}

\subsection{Datasets}
\subsubsection{Sentiment Analysis (SA).}
Our approach leverages task-specific data from a high-resource language (English) along with target language data, which we refer to as the source and target datasets, respectively. For SA, we use the following target datasets:
\begin{itemize}
\item Korean and Arabic: We use the datasets from the MTEB benchmark\footnote{\url{https://huggingface.co/datasets/mteb/multilingual-sentiment-classification}} for the binary sentiment classification task.
\item Hindi: We use the dataset from \cite{akhtar-etal-2016-hybrid}, for a three-way sentiment analysis task including \textit{positive}, \textit{negative}, and \textit{neutral} labels.
\end{itemize}
For task-specific SA instances in the source (English) languages:
\begin{itemize}
\item For Korean and Arabic, we use the SST2~\cite{socher-etal-2013-recursive} training set in English.
\item For Hindi, we construct a ternary classification dataset (SST3) from SST5\footnote{\url{https://huggingface.co/datasets/SetFit/sst5}} by merging \textit{very positive} with \textit{positive} and \textit{very negative} with \textit{negative}.
\end{itemize}
\subsubsection{Natural Language Inference (NLI).}
We construct custom target datasets for Arabic, Hindi, and Urdu by sampling instances from their respective development sets in XNLI. To simulate a low-resource setting, we partition the data into train:dev:test splits of 1500:490:500 instances, setting aside the labels in the training split.

For the Korean target dataset, we draw 5000 instances from the \textit{multi-nli-train} split of a Kaggle dataset created by translating XNLI.\footnote{\url{https://www.kaggle.com/datasets/thedevastator/korean-natural-language-inference-datasets}}

The English train split from the original XNLI~\citep{conneau2018xnlievaluatingcrosslingualsentence} benchmark serves as the source dataset for NLI tasks across all target languages. All these datasets and code to reproduce our experiments will be made publicly available upon publication.

\subsection{Model Training}
\label{sec:modeltraining}

We show results based on two different models: multilingual-e5 model~\cite{wang2024multilinguale5textembeddings} and XLM-R large~\cite{conneau2020unsupervisedcrosslingualrepresentationlearning}. Both multilingual E5 and XLM-R are trained on massive multilingual corpora, producing rich cross-lingual embeddings. Their bidirectional attention mechanisms are well-suited for capturing holistic meaning and context, unlike decoder-based models. Recent work~\cite{saattrup-nielsen-etal-2025-encoder} demonstrates that encoder models can achieve significantly better natural language understanding (NLU) performance than decoder models, often with orders of magnitude fewer parameters. Additionally, models such as XLMR offer computational efficiency and ease of fine-tuning compared to large generative language models, making them particularly suitable for experimentation and data augmentation. Accordingly, our augmentation framework leverages these encoder-based models as effective backbones, enabling high-quality and generalizable representations. Future work could explore adapting these methods to decoder-based architectures, potentially extending their benefits to a broader range of model families.

As described in Section \ref{sec:mapping}
we perform a 1:$m$ mapping of target to source text based on embedding similarity, with 
$m = 5$ as the default setting unless specified otherwise.%
\footnote{For Korean SA, we use $m = 10$.}

\begin{table*}[h!]
\centering
\begin{adjustbox}{max width=\textwidth}
\begin{tabular}{|p{0.15\linewidth}|p{0.11\linewidth}| p{0.11\linewidth}|p{0.11\linewidth}|p{0.11\linewidth}|p{0.11\linewidth}|p{0.11\linewidth}|p{0.11\linewidth}|p{0.11\linewidth}|p{0.11\linewidth}|}
\hline 
\multirow{2}{*}{\small{\textbf{Model}}} & \multicolumn{2}{|c|}{\small{\textbf{Korean}}} & \multicolumn{2}{|c|}{\small{\textbf{Arabic}}} & \multicolumn{2}{|c|}{\small{\textbf{Hindi}}}\\
\cline{2-7} 
& \small{\textbf{ko}} & \small{\textbf{en}} & \small{\textbf{ar}} & \small{\textbf{en}} & \small{\textbf{hi}} & \small{\textbf{en}}  \\
\cline{2-7} 
& \small{\textbf{dev/test}} & \small{\textbf{dev/test}} & \small{\textbf{dev/test}} & \small{\textbf{dev/test}} & \small{\textbf{dev/test}} & \small{\textbf{dev/test}}  \\
\hline 
\small{Skyline} &\small{84.47/85.23} &\small{88.76/90.61} &\small{93.95/89.09} &\small{64.91/63.92} &\small{65.81/65.48} &\small{43.35/44.37}\\ 
\hline 
\small{Baseline-PS} &\small{78.32/79.15} &\small{87.16/88.14} &\small{78.63/81.02} &\small{86.12/85.67} &\small{57.1/56.13} &\small{37.04/38.22}\\ 
\hline
\small{\ourtechnique-reg-reg} &\small{77.87/79.34} &\small{90.83/91.05} &\small{81.05/81.73} &\small{87.73/88.69} &\small{58.71/56.45} &\small{44.04/45.09}\\ 
\small{\ourtechnique-reg-lda} &\cellcolor{mygreen}\small{78.92/80.24} &\small{90.94/90.77} &\cellcolor{mygreen}\small{81.05/82.01} &\small{88.19/89.73} &\cellcolor{mygreen}\small{59.03/56.13} &\small{43.69/44.70}\\ 
\hline
\end{tabular}
\end{adjustbox}
\caption{Results for Sentiment Analysis with multilingual e5. The best numbers across all systems (excluding the Skyline) are highlighted in green.}
\label{tab:SA-me5}
\end{table*}

\begin{table*}[t]
\centering
\begin{adjustbox}{max width=\textwidth}
\begin{tabular}{|p{0.14\linewidth}|p{0.11\linewidth}| p{0.11\linewidth}|p{0.10\linewidth}|p{0.11\linewidth}|p{0.10\linewidth}|p{0.11\linewidth}|p{0.11\linewidth}|p{0.11\linewidth}|p{0.07\linewidth}|}
\hline 
\multirow{2}{*}{\small{\textbf{Model}}} & \multicolumn{2}{|c|}{\small{\textbf{Korean}}} & \multicolumn{2}{|c|}{\small{\textbf{Arabic}}} & \multicolumn{2}{|c|}{\small{\textbf{Hindi}}} & \multicolumn{2}{|c|}{\small{\textbf{Urdu}}}\\
\cline{2-9} 
& \small{\textbf{ko}} & \small{\textbf{en}} & \small{\textbf{ar}} & \small{\textbf{en}} & \small{\textbf{hi}} & \small{\textbf{en}}  & \small{\textbf{ur}} & \small{\textbf{en}}\\
\cline{2-9} 
& \small{\textbf{dev/test}}  & \small{\textbf{dev/test}}  & \small{\textbf{dev/test}}  & \small{\textbf{dev/test}}  & \small{\textbf{dev/test}} & \small{\textbf{dev/test}}  & \small{\textbf{dev/test}} & \small{\textbf{dev/test}} \\
\hline 
\small{Skyline} &\small{61.16/62.07} &\small{45.02/75.93} &\small{54.08/54} &\small{51.73/53.25} &\small{54.49/54.4} &\small{55.78/18.50} &\small{51.84/48.6} &\small{49.88/5.63}\\ 
\hline 
\small{Baseline-PS} &\small{45.98/46.17} &\small{44.86/15.57} &\small{41.63/46.4} &\small{45.38/0.04} &\small{45.10/48.2} &\small{45.66/0.10} &\small{43.06/46.8} &\small{45.26/0.66}\\ 
\hline 
\small{\ourtechnique-reg-reg} &\small{47.75/48.08} &\small{62.73/57.56} &\small{47.14/49.6} &\small{51.65/34.35} &\small{52.04/51} &\small{53.13/27.42} &\small{45.71/49.4} &\small{53.57/23.83}\\ 
\small{\ourtechnique-reg-lda} &\cellcolor{mygreen}\small{50.96/52} &\small{57.27/54.79} &\cellcolor{mygreen}\small{47.14/50.2} &\small{52.13/38.84} &\cellcolor{mygreen}\small{53.27/52} &\small{54.14/38.06} &\cellcolor{mygreen}\small{49.39/49} &\small{54.22/33.81}\\ 
\hline
\end{tabular}
\end{adjustbox}
\caption{Results for NLI with multilingual e5. The best numbers across all systems (excluding the Skyline) are highlighted in green.}
\label{tab:NLI-me5}
\end{table*}

\subsubsection{Skyline.}
In this system, we include the actual target labels during training. Only the target text and corresponding ground-truth labels are used during finetuning; this allows us to establish an upper-bound on target task performance. The skyline is meant for the target language test sets. However, we also compute accuracies on English test instances using this data as a measure of forgetting. We note that the scores on English are based on the best-performing model for the target text\footnote{Validation data in the target language is used to select the best checkpoint.} instead of English.

\subsubsection{Baseline.}
We setup the following baselines for comparison:
\begin{itemize}
\item Baseline-PS (project-source): We simply project the labels from the source text onto the mapped target text and train the model for the given task on the target text. This baseline is reported for both SA and NLI tasks, while the next two baselines are shown only for SA, as they require translating the English source corpus—a process that is prohibitively time-consuming for the large-scale XNLI dataset. All references to the \textit{baseline} throughout this work corresponds to the Baseline-PS setup.
\item Baseline-TT: This is the translate-train baseline, where the model is trained on the entire source dataset\footnote{SST2 train for Korean and Arabic SA, and SST3 train for Hindi SA}, translated into the target language. While effective, this approach is computationally expensive due to the large size of the source dataset, requiring significant time both for translation and training. 
\item Baseline-KD: This system is based on knowledge distillation and uses the following three steps. First, we start with the translate train system in Baseline-TT and use it to pseudolabel the target text. The pseudolabeled target text is then used to train another (student) model which is subsequently evaluated.\footnote{Pretrained XLMR-large is used to train both the teacher and the student.}
It performs way worse than all other evaluated systems. This could be attributed to the fact that pseudolabels generated using this method match the actual labels of target text only 15.16\%, 14.55\% and 13.15\% of the time for Korean, Arabic and Hindi respectively.
\end{itemize}

\subsubsection{\ourtechnique-\texttt{reg-reg} (Interpolation).}
In this system, we use the projected source labels as in the Baseline-PS. However, instead of using only the target text, we interpolate the source and mapped target text by simply adding their respective embeddings. The resulting combined embedding is then passed to the classifier head for task-specific fine-tuning.

\subsubsection{\ourtechnique-\texttt{reg-lda} (Projection then Interpolation).}
This system builds on the previous interpolation approach but incorporates an additional projection step. Before adding the target text embedding to the source embedding, we first project the target embedding onto a basis that captures language separability. To compute this basis, we apply the method described in Section \ref{sec:deriv-basis} using data from the target language and English, with varying values of $k$ (number of dimensions) and $n$ 
(number of sentences used). The final scores are reported using the hyperparameter combination that yields the best performance on the validation set. More details about the hyperparameters are provided in supplementary material.

Tables \ref{tab:SA-me5} and \ref{tab:NLI-me5} compare the performance of four systems across target languages using the me5 model. We observe consistent improvements over Baseline-PS after applying interpolation. Notably, the system with projection before interpolation (\ourtechnique-\texttt{reg-lda}) outperforms the simpler interpolation method (\ourtechnique-\texttt{reg-reg}). 
Similarly, Tables \ref{tab:SA-xlmr-l} and \ref{tab:NLI-xlmr-l} present scores using embeddings derived from the XLM-R large model. 

For XLM-R large on the SA task, Baseline-TT achieves the highest performance; however, it relies on extensive training over a large machine-translated dataset. This approach is not only computationally expensive and time-consuming -- especially when the source corpus is large -- but also less feasible for low-resource target languages due to potential issues related to translation quality. For this reason, we do not report Baseline-TT or Baseline-KD results for the NLI task, where the source English dataset (MNLI) consists of approximately 433K training pairs.

\begin{table*}[h!]
\centering
\begin{adjustbox}{max width=\textwidth}
\begin{tabular}{|p{0.17\linewidth}|p{0.11\linewidth}| p{0.11\linewidth}|p{0.11\linewidth}|p{0.11\linewidth}|p{0.11\linewidth}|p{0.11\linewidth}|p{0.11\linewidth}|p{0.11\linewidth}|p{0.11\linewidth}|}
\hline 
\multirow{2}{*}{\small{\textbf{Model}}} & \multicolumn{2}{|c|}{\small{\textbf{Korean}}} & \multicolumn{2}{|c|}{\small{\textbf{Arabic}}} & \multicolumn{2}{|c|}{\small{\textbf{Hindi}}}\\
\cline{2-7} 
& \small{\textbf{ko}} & \small{\textbf{en}} & \small{\textbf{ar}} & \small{\textbf{en}} & \small{\textbf{hi}} & \small{\textbf{en}}  \\
\cline{2-7} 
& \small{\textbf{dev/test}} & \small{\textbf{dev/test}} & \small{\textbf{dev/test}} & \small{\textbf{dev/test}} & \small{\textbf{dev/test}} & \small{\textbf{dev/test}}  \\
\hline 
\small{Skyline} &\small{89.5/88.56} &\small{87.96/89.02} &\small{97.58/91.5} &\small{82.45/84.79}  &\small{69.03/70.97} &\small{42.32/43}\\ 
\hline 
\small{Baseline-PS} &\small{73.14/71.17} &\small{60.78/60.35} &\small{71.77/66.86} &\small{51.03/49.97} &\small{45.81/45.16} &\small{15.6/16.75}\\ 
\hline 
\small{Baseline-TT} &\cellcolor{mygreen}\small{85.45/86.20} &\small{93.23/93.96} &\cellcolor{mygreen}\small{85.48/84.84} &\small{93.81/94.51} &\cellcolor{mygreen}\small{62.58/59.68} &\small{42.89/44.43}\\ 
\hline
\small{Baseline-KD} &\small{16.05/13.91} &\small{12.16/10.87} &\small{16.94/17.42} &\small{32.45/33.44} &\small{33.87/30.65} &\small{46.90/48.54}\\ 
\hline 
\small{\ourtechnique-reg-reg} &\small{82.22/83.69} &\small{92.32/93.41} &\small{83.87/82.86} &\small{91.63/92.81} &\small{52.58/50.65} &\small{41.63/42.06}\\ 
\small{\ourtechnique-reg-lda} &\cellcolor{lightgreen}\small{83.12/83.58} &\small{91.28/91.87} &\cellcolor{lightgreen}\small{85.08/85.55} &\small{91.4/93.08} &\cellcolor{lightgreen}\small{62.26/57.42} &\small{44.04/45.36}\\ 
\hline
\end{tabular}
\end{adjustbox}
\caption{Results for SA using XLMR-large. The best and second-best numbers (excluding the Skyline) are highlighted in green and light green, respectively.}
\label{tab:SA-xlmr-l}
\end{table*}

\begin{table*}[h!]
\centering
\begin{adjustbox}{max width=\textwidth}
\begin{tabular}{|p{0.14\linewidth}|p{0.11\linewidth}| p{0.11\linewidth}|p{0.10\linewidth}|p{0.11\linewidth}|p{0.10\linewidth}|p{0.11\linewidth}|p{0.11\linewidth}|p{0.11\linewidth}|p{0.07\linewidth}|}
\hline 
\multirow{2}{*}{\small{\textbf{Model}}} & \multicolumn{2}{|c|}{\small{\textbf{Korean}}} & \multicolumn{2}{|c|}{\small{\textbf{Arabic}}} & \multicolumn{2}{|c|}{\small{\textbf{Hindi}}} & \multicolumn{2}{|c|}{\small{\textbf{Urdu}}}\\
\cline{2-9} 
& \small{\textbf{ko}} & \small{\textbf{en}} & \small{\textbf{ar}} & \small{\textbf{en}} & \small{\textbf{hi}} & \small{\textbf{en}}  & \small{\textbf{ur}} & \small{\textbf{en}}\\
\cline{2-9} 
& \small{\textbf{dev/test}}  & \small{\textbf{dev/test}}  & \small{\textbf{dev/test}}  & \small{\textbf{dev/test}}  & \small{\textbf{dev/test}} & \small{\textbf{dev/test}}  & \small{\textbf{dev/test}} & \small{\textbf{dev/test}} \\
\hline 
\small{Skyline} &\small{76.02/75.93} &\small{77.75/34.69} &\small{63.06/62} &\small{73.53/44.07} &\small{66.12/64} &\small{76.91/39.54} &\small{62.45/61.2} &\small{68.11/33.47}\\ 
\hline 
\small{Baseline-PS} &\small{36.67/36.58} &\small{37.55/71.22} &\small{34.9/30.6} &\small{33.33/100} &\small{35.31/33} &\small{33.78/97.31} &\small{35.10/31.6} &\small{34.5/96.43}\\ 
\hline 
\small{\ourtechnique-reg-reg} &\small{50.08/50.28} &\small{71.81/39.48} &\small{58.16/57} &\small{66.47/42.65} &\small{42.45/40.2} &\small{55.46/51.06} &\small{38.16/34.4} &\small{53.45/49.92}\\ 
\small{\ourtechnique-reg-lda} &\cellcolor{mygreen}\small{66.43/65.35} &\small{83.09/32.83} &\cellcolor{mygreen}\small{60.41/62.2} &\small{68.35/38.10} &\cellcolor{mygreen}\small{51.02/49.4} &\small{57.35/55.35} &\cellcolor{mygreen}\small{51.22/47} &\small{70.6/36.95}\\ 
\hline
\end{tabular}
\end{adjustbox}
\caption{Results for NLI using XLMR-large. The best numbers across all systems (excluding the Skyline) are highlighted in green.}
\label{tab:NLI-xlmr-l}
\end{table*}

\section{Discussion}

\subsection{Performance with XLMR-large}
While the skyline scores improve compared to me5 (compare Skylines scores in Tables \ref{tab:SA-me5} and \ref{tab:SA-xlmr-l}, similarly in Tables \ref{tab:NLI-me5} and \ref{tab:NLI-xlmr-l}), Baseline-PS scores for XLMR-large drop.
More importantly, the gains from interpolation become significantly more pronounced, as highlighted in Tables \ref{tab:SA-xlmr-l} and \ref{tab:NLI-xlmr-l}, showing an approximately 24\% improvement for Arabic SA and a 22\% boost for Hindi NLI, compared to the respective Baseline-PS. Except for Hindi and Urdu NLI,  interpolation using XLM-R large achieves the best performance across all other tasks and languages.

To better understand why Baseline-PS scores are higher with me5, we also report the label accuracy of the train sets for both models across the two tasks in Table \ref{tab:label-acc} used in the Baseline-PS (and, our \ourtechnique-based systems). Higher Baseline-PS accuracy using me5 can be attributed to the contrastive pretraining and supervised finetuning of me5 which is more suited for retrieval tasks and semantic search. Note that SA for Korean and Arabic is a binary classification task, while SA for Hindi and NLI for all four languages involve three-way classification. Interestingly, despite starting with projected labels that are only marginally better than random (see Tables \ref{tab:SA-xlmr-l} and \ref{tab:NLI-xlmr-l}), the interpolation approach effectively leverages information from both source and target embeddings, resulting in substantial performance gains across languages.


\begin{table*}[h!]
\centering
\begin{adjustbox}{max width=\textwidth}
\begin{tabular}{|p{0.2\linewidth}|p{0.12\linewidth}|p{0.2\linewidth}|p{0.2\linewidth}|p{0.2\linewidth}|}
\hline 
& \multicolumn{2}{c|}{\small{\textbf{XLMR-large}}} & \multicolumn{2}{c|}{\small{\textbf{me5}}}\\ 
\hline 
\multirow{2}{*}{\small{\textbf{SA}}}& \multicolumn{2}{c|}{\small{\textbf{Baseline (projected labels)}}} & \multicolumn{2}{c|}{\small{\textbf{Baseline (projected labels)}}}\\ 
\cline{2-5}
& \small{\textbf{top-1}} & \small{\textbf{top-k}}  & \small{\textbf{top-1}} & \small{\textbf{top-k}} \\ 
\hline 
\small{Korean} & \small{56.62} & \small{56.54} & \small{68.98}  & \small{69.47}\\
\small{Arabic} & \small{ 55.31} & \small{55.86} & \small{67.83} & \small{68.08} \\
\small{Hindi} & \small{39.92} & \small{39.07} & \small{48.47} & \small{49.73} \\
\hline 
\multirow{2}{*}{\small{\textbf{NLI}}}& \multicolumn{2}{c|}{\small{\textbf{Baseline (projected labels)}}} & \multicolumn{2}{c|}{\small{\textbf{Baseline (projected labels)}}} \\ 
\cline{2-5}
& \small{\textbf{top-1}} & \small{\textbf{top-k}} & \small{\textbf{top-1}} & \small{\textbf{top-k}}\\ 
\hline 
\small{Korean} & \small{34.48} & \small{35.76} & \small{20.66}  & \small{39.15}\\
\small{Arabic} & \small{35.00} & \small{35.23} & \small{39.40}  & \small{39.09}\\
\small{Hindi} & \small{33.93 } & \small{34.92} & \small{40.80}  & \small{40.68}\\
\small{Urdu} & \small{36.07} & \small{34.61} & \small{39.67}  & \small{39.35}\\
\hline 
\end{tabular}
\end{adjustbox}
\caption{Label projection accuracies for both SA and NLI.}
\label{tab:label-acc}
\end{table*}







\begin{table}[h!]
\centering
\begin{adjustbox}{max width=\textwidth}
\begin{tabular}{|p{0.2\linewidth}|p{0.12\linewidth}|p{0.12\linewidth}|}
\hline 
& \multicolumn{2}{c|}{\small{\textbf{With LABSE-based mapping}}} \\ 
\hline 
\multirow{2}{*}{\small{\textbf{SA}}}& \multicolumn{2}{c|}{\small{\textbf{Baseline (projected labels)}}} \\ 
\cline{2-3}
& \small{\textbf{top-1}} & \small{\textbf{top-k}}   \\ 
\hline 
\small{Korean} & \small{63.24} & \small{62.72} \\
\small{Arabic} & \small{62.52} & \small{62.46}  \\
\small{Hindi} & \small{41.72} & \small{42.58}  \\
\hline 
\multirow{2}{*}{\small{\textbf{NLI}}}& \multicolumn{2}{c|}{\small{\textbf{Baseline (projected labels)}}} \\ 
\cline{2-3}
& \small{\textbf{top-1}} & \small{\textbf{top-k}} \\ 
\hline 
\small{Korean} & \small{5.62} & \small{36.32}\\
\small{Arabic} & \small{37.67} & \small{37.92} \\
\small{Hindi} & \small{34.93} & \small{37.16} \\
\small{Urdu} & \small{37.20} & \small{38.59} \\
\hline 
\end{tabular}
\end{adjustbox}
\caption{Label projection accuracies for both SA and NLI with LABSE mapping.}
\label{tab:label-acc-LABSE}
\end{table}


\subsection{Minimal Forgetting!}
We observe an interesting consequence of using \ourtechnique: the performance on English evaluation sets also improves compared to Baseline-PS (see Tables \ref{tab:SA-me5}, \ref{tab:NLI-me5}, \ref{tab:SA-xlmr-l} and \ref{tab:NLI-xlmr-l}). This can likely be attributed to the model's exposure to source text during interpolation-based training for the NLU tasks, which serves as an implicit regularizer. By reinforcing the learned representations, this exposure mitigates catastrophic forgetting and helps the model retain its performance on the source language. The English dev and test sets are both used as test sets, and the reported scores correspond to the best-performing models selected based on the respective target dev set. For SA, we evaluate using the English test sets from the multilingual sentiment classification dataset in MTEB\footnote{https://huggingface.co/datasets/mteb/multilingual-sentiment-classification} across all target languages, while for NLI, we report results using the original dev and test splits from XNLI.


\begin{table*}[h!]
\centering
\begin{adjustbox}{max width=\textwidth}
\begin{tabular}{|p{0.17\linewidth}|p{0.11\linewidth}| p{0.11\linewidth}|p{0.11\linewidth}|p{0.11\linewidth}|p{0.11\linewidth}|p{0.11\linewidth}|p{0.11\linewidth}|p{0.11\linewidth}|p{0.11\linewidth}|}
\hline 
\multirow{2}{*}{\small{\textbf{Model}}} & \multicolumn{2}{|c|}{\small{\textbf{Korean}}} & \multicolumn{2}{|c|}{\small{\textbf{Arabic}}} & \multicolumn{2}{|c|}{\small{\textbf{Hindi}}}\\
\cline{2-7} 
& \small{\textbf{ko}} & \small{\textbf{en}} & \small{\textbf{ar}} & \small{\textbf{en}} & \small{\textbf{hi}} & \small{\textbf{en}}  \\
\cline{2-7} 
& \small{\textbf{dev/test}} & \small{\textbf{dev/test}} & \small{\textbf{dev/test}} & \small{\textbf{dev/test}} & \small{\textbf{dev/test}} & \small{\textbf{dev/test}}  \\
\hline 
\small{Baseline-LABSE} &\small{79.97/81.25} &\small{87.84/88.69}  &\small{83.87/82.15} &\small{88.65/76.27} &\small{58.06/55.81} &\small{44.84/46.18}\\
\hline
\small{\ourtechnique-LABSE} &\small{82.90/83.69} &\small{91.06/92.59} &\small{83.87/86.83} &\small{88.65/89.84} &\small{63.55/57.10} &\small{43.69/44.32}\\ 
\hline

\end{tabular}
\end{adjustbox}
\caption{Results for SA using XLMR-large with LABSE based mapping.}
\label{tab:SA-xlmr-l-labse}
\end{table*}

\begin{table*}[h!]
\centering
\begin{adjustbox}{max width=\textwidth}
\begin{tabular}{|p{0.14\linewidth}|p{0.11\linewidth}| p{0.11\linewidth}|p{0.10\linewidth}|p{0.11\linewidth}|p{0.10\linewidth}|p{0.11\linewidth}|p{0.11\linewidth}|p{0.11\linewidth}|p{0.07\linewidth}|}
\hline 
\multirow{2}{*}{\small{\textbf{Model}}} & \multicolumn{2}{|c|}{\small{\textbf{Korean}}} & \multicolumn{2}{|c|}{\small{\textbf{Arabic}}} & \multicolumn{2}{|c|}{\small{\textbf{Hindi}}} & \multicolumn{2}{|c|}{\small{\textbf{Urdu}}}\\
\cline{2-9} 
& \small{\textbf{ko}} & \small{\textbf{en}} & \small{\textbf{ar}} & \small{\textbf{en}} & \small{\textbf{hi}} & \small{\textbf{en}}  & \small{\textbf{ur}} & \small{\textbf{en}}\\
\cline{2-9} 
\hline
\small{Baseline-LABSE} &\small{42.45/43.8} &\small{47.43/11.06} &\small{42.45/43.8} &\small{47.43/11.06} &\small{42.24/46.2} &\small{47.23/0} &\small{41.43/43.4} &\small{41.53/3.89}\\ 
\hline
\small{\ourtechnique-LABSE} &\small{67.55/66.8} &\small{76.27/37.58} &\small{67.55/66.8} &\small{76.26/37.58} &\small{66.53/68.2} &\small{79.56/37.88} &\small{63.67/63.2} &\small{77.71/35.95}\\ 
\hline
\end{tabular}
\end{adjustbox}
\caption{Results for NLI using XLMR-large with LABSE-based mapping.}
\label{tab:NLI-xlmr-l-labse}
\end{table*}

\subsection{Target-to-source mapping}
Since XLM-R is not specifically trained for producing sentence-level embeddings, we investigate whether using LaBSE embeddings for target-to-source instance mapping can yield better results. Tables~\ref{tab:SA-xlmr-l-labse} and~\ref{tab:NLI-xlmr-l-labse} lists the baseline and interpolation results\footnote{\ourtechnique-reg-reg is implemented for this.} for XLM-R when the mapping is derived using LaBSE sentence embeddings, based on CLS token cosine similarity. 

We observe that the Baseline-LABSE significantly outperforms the default Baseline, where both mapping and fine-tuning uses XLM-R. Moreover, interpolation based on the LaBSE-derived mapping further improves scores over both baselines. We also report label accuracies under the LABSE mapping in Table~\ref{tab:label-acc-LABSE}, which are generally improved compared to those in Table~5—particularly for the SA task.

Importantly, we emphasize that our proposed method continues to be effective even when the label alignment is imperfect. Top-k instance mappings allow the model to learn sentiment or relational cues by implicitly disentangling various semantic aspects of the sentence or sentence pair. Through interpolation of target embeddings with semantically related source embeddings—even when the source and target labels differ—the model is able to capture richer representations learning nuanced classification boundaries.

\subsubsection{Random target to source mapping strategy.}
As an ablation for similarity-based mapping, we also experiment with random mappings of target to source text for Hindi sentiment analysis. This results in a drop in baseline accuracy (with label projection) from $45.16$ to $39.35$. Similarly, the interpolation method (\ourtechnique-\texttt{reg-reg}) sees a sharp decline from $50.65$ to $39.68$ using XLM-R large. This is clear evidence of the need for a similarity-based mapping to choose source text, corresponding to unlabeled text in the target language. 

\subsection{Impact of higher mapping value $m$}
Table~\ref{tab:mapping-trend} shows the baseline performance with values of $m \in [1,3,5,7,10]$ for Arabic SA. We find that the scores keep increasing until 7 after which performance unsurprisingly begins to degrade; increasing $m$ beyond a point would compromise the degree of similarity between source and target text and subsequently the projected labels would not be as accurate. We also show skyline and interpolation numbers for $m=7$. With our default choice of $m=5$, we see significant improvements with interpolation especially with \ourtechnique-\texttt{reg-lda} yielding a 17.27\% improvement compared to the Baseline. 
(Scores with varying the choices of two hyperparameters for our LDA-based method are shared in the supplementary material.) 

\begin{table}[h!]
\centering
\begin{adjustbox}{max width=\textwidth}
\begin{tabular}{|p{0.3\linewidth}|p{0.2\linewidth}| p{0.2\linewidth}|}
\hline 
\multirow{2}{*}{\small{\textbf{Model}}} & \multicolumn{2}{|c|}{\small{\textbf{Arabic}}} \\
\cline{2-3} 
& \small{\textbf{ar}} & \small{\textbf{en}} \\
& \small{\textbf{dev/test}}  & \small{\textbf{dev/test}}   \\
\hline 
\multicolumn{3}{|l|}{\small{}}\\
\hline
\small{m=1} & \small{52.42/49.72}  & \small{50.92/49.92} \\
\small{m=3} & \small{52.82/50.14}  & \small{50.92/49.92} \\
\small{m=5} & \small{71.77/66.86}  & \small{51.03/49.97} \\
\small{m=7} & \cellcolor{mygreen}\small{75.81/70.54}  & \small{50.92/49.92} \\
\small{m=10} & \small{72.18/66.57}  & \small{50.8/49.92} \\
\small{m=15} & \small{71.37/67.14}  & \small{51.49/50.47} \\
\hline 
\multicolumn{3}{|c|}{\small{with m=7}}\\
\hline 
\small{Skyline} & \small{97.98/91.50}  & \small{84.29/84.51} \\
\small{Baseline} & \small{75.81/70.54}  & \small{50.92/49.92} \\
\small{\ourtechnique-reg-reg} & \small{77.82/71.67}  & \small{91.28/92.7} \\
\small{\ourtechnique-reg-lda} & \small{81.85/82.72}  & \small{88.19/88.85} \\
\hline 
\end{tabular}
\end{adjustbox}
\caption{\label{tab:mapping-trend}Varying $m$ in top-$m$ source to target text mapping.}
\end{table}

\section{Conclusion}
In this study, we introduce an embedding-based interpolation framework \ourtechnique designed to enhance cross-lingual transfer in multilingual language models. By blending source and target language embeddings, our approach aims to bridge representation gaps, facilitating more effective transfer of semantic information across languages.

Our empirical evaluations on tasks such as sentiment analysis and natural language inference across multiple target languages, including Hindi, Arabic, and Korean, demonstrate that interpolation-based methods consistently outperform standard cross-lingual baselines. Notably, our projection-based interpolation strategy, which leverages language-separability subspaces, effectively reduces linguistic interference while preserving essential semantic features.

These findings underscore the potential of embedding interpolation as a viable technique for mitigating representation discrepancies in multilingual contexts. By seamlessly integrating information from both source and target languages, our method offers a promising avenue for improving performance in cross-lingual natural language understanding tasks, particularly in scenarios involving low-resource languages.

Future research could explore the integration of interpolation techniques with other alignment methods, such as adversarial training or optimal transport alignment, to further enhance cross-lingual transfer. Additionally, investigating the application of embedding interpolation in encoder-decoder or decoder-only models may provide insights into balancing language-neutral and language-specific information, as highlighted in recent surveys on cross-lingual alignment.~\citep{hammerl-etal-2024-understanding}

\newpage
\bibliography{aaai2026}

\end{document}